\documentclass[a4paper,10pt,twocolumn]{article}
\usepackage[T1]{fontenc}
\usepackage[utf8]{inputenc}
\usepackage{newtxtext,newtxmath}  % Modern Times fonts with all sizes

\usepackage[
  a4paper,
  top=25mm, bottom=25mm,
  left=18mm, right=18mm,
  headheight=0pt, headsep=0pt,
  footskip=10mm
]{geometry}

\usepackage{amsmath,amsfonts}

\usepackage{graphicx}
\usepackage{booktabs}
\usepackage{threeparttable}
\usepackage{caption}
\usepackage{tikz}
\usetikzlibrary{shapes.geometric, arrows.meta, positioning,
                fit, backgrounds, calc}

\usepackage{xcolor}
\usepackage{gensymb}
\usepackage{colortbl}
\usepackage{textcomp}
\usepackage{csquotes}
\usepackage{dblfloatfix}
\usepackage{placeins}
\usepackage{ragged2e}  % for \justifying command

\usepackage{cite}
\usepackage{hyperref}
\hypersetup{
  colorlinks=false,
  pdfborder={0 0 0}
}

\usepackage{titlesec}
\titleformat{\section}
  {\normalfont\bfseries\fontsize{11}{13}\selectfont\centering}
  {\thesection.}{0.5em}{}
\titleformat{\subsection}
  {\normalfont\bfseries\fontsize{10}{12}\selectfont}
  {\thesubsection}{0.5em}{}
\titlespacing*{\section}   {0pt}{10pt}{6pt}
\titlespacing*{\subsection}{0pt}{8pt}{4pt}

\usepackage[bottom]{footmisc}
\renewcommand{\footnoterule}{%
  \kern-3pt\hrule width\columnwidth height 0.4pt\kern3pt}

\newlength{\abstractwidth}
\begin{document}
\pagestyle{empty}

%% ════════════════════════════════════════════════════════════════
%%  FULL-WIDTH TITLE BLOCK
%% ════════════════════════════════════════════════════════════════
\twocolumn[{%
  \begin{minipage}{\textwidth}

    %% ── Title: 16pt bold, centered ─────────────────────────────
    \begin{center}
      {\fontsize{16}{18}\selectfont
        Automotive Engineering-Centric Agentic AI Workflow Framework 
      \par}

      \vspace{2pt}

      {\fontsize{12}{13}\selectfont
      %Explainable Workflows for CAE Mode Classification and CFD Field Prediction
      \par}

      \vspace{12pt}

      {\fontsize{11}{15}\selectfont\bfseries
        Tong Duy Son, Zhihao Liu, Piero Brigida, Yerlan Akhmetov,  \\
        Gurudevan Devarajan, Kai Liu, Ajinkya Bhave
      }
    \end{center}

    \vspace{5pt}

    %% ════════════════════════════════════════════════════════════
    %%  ABSTRACT 
    %% ════════════════════════════════════════════════════════════
    \noindent\hfill
    \begin{minipage}{\abstractwidth}
      \setlength{\parindent}{4mm}
      \setlength{\parskip}{0pt}
      {\fontsize{9}{11}\selectfont\justifying
      \noindent\hspace{4mm}%

\begin{abstract}
Engineering workflows such as design optimization, simulation-based diagnosis, control tuning, 
and model-based systems engineering (MBSE) are iterative, constraint-driven, and shaped by prior decisions. 
Yet many AI methods still treat these activities as isolated tasks rather than as parts of a broader workflow. 
This paper presents Agentic Engineering Intelligence (AEI), an industrial vision framework that models engineering 
workflows as constrained, history-aware sequential decision processes in which AI agents support 
engineer-supervised interventions over engineering toolchains. 
AEI links an offline phase for engineering data processing and workflow-memory construction with an online 
phase for workflow-state estimation, retrieval, and decision support. 
A control-theoretic interpretation is also possible, in which engineering objectives act as reference signals, 
agents act as workflow controllers, and toolchains provide feedback for intervention selection. 
Representative automotive use cases in suspension design, reinforcement learning tuning, multimodal 
engineering knowledge reuse, aerodynamic exploration, and MBSE show how diverse workflows can be 
expressed within a common formulation. 
Overall, the paper positions engineering AI as a problem of process-level intelligence and outlines 
a practical roadmap for future empirical validation in industrial settings.
\end{abstract}

      \vspace{5pt}
      }
    \end{minipage}%
    \hfill\null

    \vspace{20pt}
  \end{minipage}
}]%% ── end \twocolumn[...] ──────────────────────────────────────

%% ── Footnotes for affiliations (appear at bottom) ─────────────
{\renewcommand{\thefootnote}{}\footnotetext{The authors are with Siemens Digital Industries Software, Leuven, Belgium. Contact Email: son.tong@siemens.com.  
This work is part of the ROBUSTIFAI project (grant agreement No.
101212818) funded by Horizon Europe – the Framework Programme for Research
and Innovation. The work also benefited from the Flanders Innovation \&
Entrepreneurship VLAIO funded project SATISFY.AI.}}

%% ════════════════════════════════════════════════════════════════
%%  TWO-COLUMN BODY
%% ════════════════════════════════════════════════════════════════

\section{Introduction}

\begin{figure*}[!t]
\centering
\includegraphics[width=\textwidth]{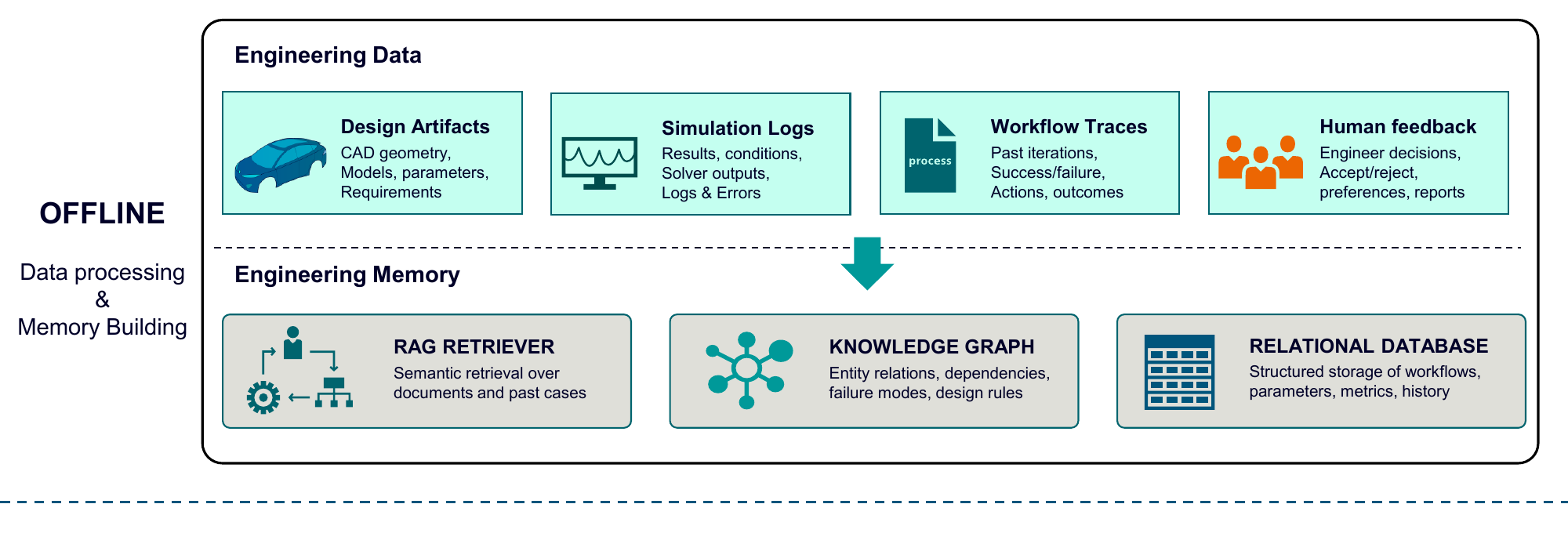}
\includegraphics[width=\textwidth]{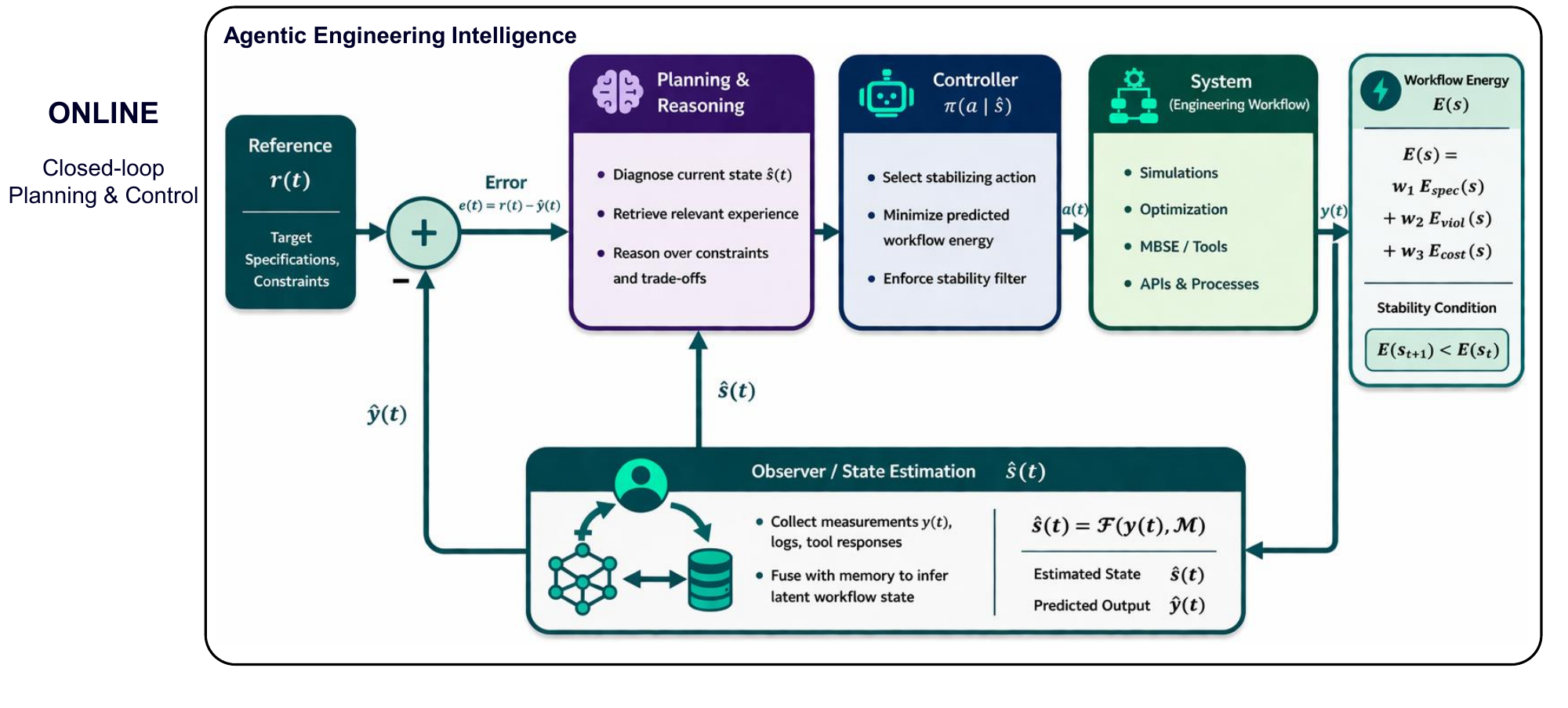}
\caption{The Agentic Engineering Intelligence (AEI) framework consists of two coupled phases.
Top: an offline engineering data processing and memory-building phase, in which design artifacts, simulation logs,
workflow traces, and human feedback are structured into reusable engineering memory, including retrieval stores,
knowledge graphs, and relational workflow records.
Bottom: an online closed-loop planning and control phase, in which the agent estimates workflow state, retrieves relevant prior experience, reasons over constraints and trade-offs, selects actions, and evaluates them through a stability-inspired criterion.
The use cases demonstrate different parts of this common architecture. The formulation section clarifies the shared state, memory, constraint, and action abstractions; multimodal engineering data focuses on offline memory construction; suspension design and RL tuning emphasize online observation, reasoning, and control adjustment; aerodynamic design couples offline surrogate knowledge with online design exploration; and MBSE illustrates toolchain level reasoning over structured engineering workflows.}
\label{fig:learning_phases}
\end{figure*}

Engineering design and development workflows, such as suspension system design, reinforcement learning (RL) control tuning, external aerodynamic design, 
and model-based systems engineering (MBSE), are inherently iterative, multi-objective, and constraint-driven.
Progress toward feasible and high performance solutions typically emerges through repeated cycles of simulation, evaluation, diagnosis, and modification.
These workflows require engineers to balance conflicting requirements, interpret heterogeneous outputs such as logs and performance metrics, and make decisions under uncertainty~\cite{vanderauweraer2007structural}.

Despite recent advances in simulation, optimization, and AI, the overall structure of engineering workflows still often treat engineering problems as 
isolated tasks, such as parameter optimization, surrogate modeling, or prediction, rather than as parts of a broader sequential decision making process.
This limits their ability to support the adaptive, history-aware, and intervention driven character of real engineering practice~\cite{miller2018sdp}.
Several challenges follow from this mismatch.

First, engineering workflows continue to rely heavily on expert intuition and experience.
When optimization fails or generates infeasible solutions, diagnosis often depends on manual inspection of logs, interpretation of constraint violations, 
and reasoning about complex interactions among design variables.
Many workflows are also only partially observable: root causes such as incompatible constraints, latent couplings, or model inadequacies are not 
directly visible from outputs alone.
Engineers therefore infer hidden relationships through repeated experimentation, leading to costly trial-and-error cycles.

Second, historical engineering data are rarely used in a systematic way.
Large volumes of workflow data are routinely generated, including simulation outputs, boundary conditions, design iterations, intervention histories, 
and failure cases.
Yet these records are often weakly structured and underutilized, so similar failure modes are repeatedly encountered while valuable institutional 
knowledge remains implicit.
This challenge is closely related to the longstanding need for traceability across evolving engineering environments, 
where requirements, models, and implementation artifacts must remain connected over extended development cycles~\cite{lu2022mbseontology}. 

Third, conventional optimization pipelines are usually organized around fixed objectives and constraints, with limited support for alternative interventions such as revising constraints, reformulating objectives, modifying models, or redirecting the search process.
When these pipelines fail, they often provide limited guidance on what action should come next.

These observations motivate a shift in perspective: from optimizing individual engineering artifacts to modeling and improving the engineering process itself.
Under this view, simulation, optimization, diagnosis, and design iterations are treated not as isolated endpoints, but as interconnected elements of a closed-loop decision system.
This perspective motivates \emph{Agentic Engineering Intelligence} (AEI), a framework in which engineering workflows are modeled as constrained, history-aware sequential decision processes 
and AI agents are positioned as controllers operating over engineering toolchains.
Recent literature on LLM-based autonomous agents highlights the importance of memory, planning, and tool use in multi-step problem solving~\cite{wang2025llmagents}.
AEI adopts these ideas in an engineering setting, but places stronger emphasis on constraint handling, workflow traceability, 
and engineer-supervised intervention over domain specific toolchains.

This perspective also admits a useful control-theoretic interpretation.
Engineering objectives may be viewed as reference signals, AI agents as controllers, and engineering toolchains as the plant through 
which interventions are executed and evaluated.
Rather than relying solely on a predefined optimization pipeline, the agent is treated as an adaptive decision maker that observes workflow state, 
reasons over constraints and historical context, and selects actions intended to move the process toward feasible and desirable outcomes.
To guide action selection in these partially observed settings, AEI introduces a stability-inspired workflow-energy view in which progress is estimated 
from three practical ingredients: performance gap, constraint violation, and workflow cost.
This energy is used as a practical guidance signal rather than as a formal proof of stability.
Offline learning from prior trajectories, preference evaluation, and memory-augmented reasoning then provide a basis for 
extracting reusable decision structure from prior engineering activity and transferring it across engineering domains~\cite{levine2020offline}.

Figure~\ref{fig:learning_phases} provides a unifying view of the use cases discussed in this paper.
The multimodal engineering data section is concerned primarily with the offline phase, where heterogeneous documents and recordings are 
transformed into searchable engineering memory.
The suspension, RL control tuning, and aerodynamic design sections emphasize the online phase, 
in which the agent observes workflow state, retrieves relevant prior experience, and proposes interventions under performance and constraint considerations.
The MBSE section spans both phases by linking structured historical workflow information with ongoing reasoning and action over engineering toolchains.
The use cases should therefore be read not as isolated applications, but as different cross-sections through the same AEI architecture.

This paper is framed as an industrial vision framework rather than as a claim of fully validated autonomous engineering control.
Its goal is to define a common abstraction, a compact formulation, and representative workflow patterns that can guide future implementation and empirical evaluation across engineering domains.
Its practical value will depend on retrieval quality, domain specific tool integration, and effective human oversight in agent decisions.

The main contributions of this work are summarized as follows:

\begin{enumerate}
  \setlength{\itemsep}{0pt}
  \setlength{\parsep}{0pt}
    \item Proposing AEI as an industrial vision framework that recasts engineering workflows as constrained, history-aware sequential decision processes, shifting the focus from artifact-level optimization to process-level intelligence.
    \item Providing a compact formulation of workflow state, observations, memory retrieval, constraints, interventions, and utility for engineering-agent reasoning over toolchains.
    \item Interpreting agentic AI systems through a control-theoretic lens and introducing a practical workflow-energy heuristic for ranking candidate interventions using performance gaps, constraint violations, and workflow cost.
    \item Illustrating the framework across multiple automotive engineering workflows, including suspension design, RL control tuning, multimodal engineering memory construction, aerodynamic design exploration, and MBSE integration.
\end{enumerate}

The next section presents a compact formulation of AEI, after which representative automotive workflows show how the framework organizes reasoning, memory, human oversight, and tool interaction within a common engineering AI architecture.

\FloatBarrier
\section{AEI Formulation}

\begin{figure*}[!t]
\centering
\includegraphics[width=0.75\textwidth]{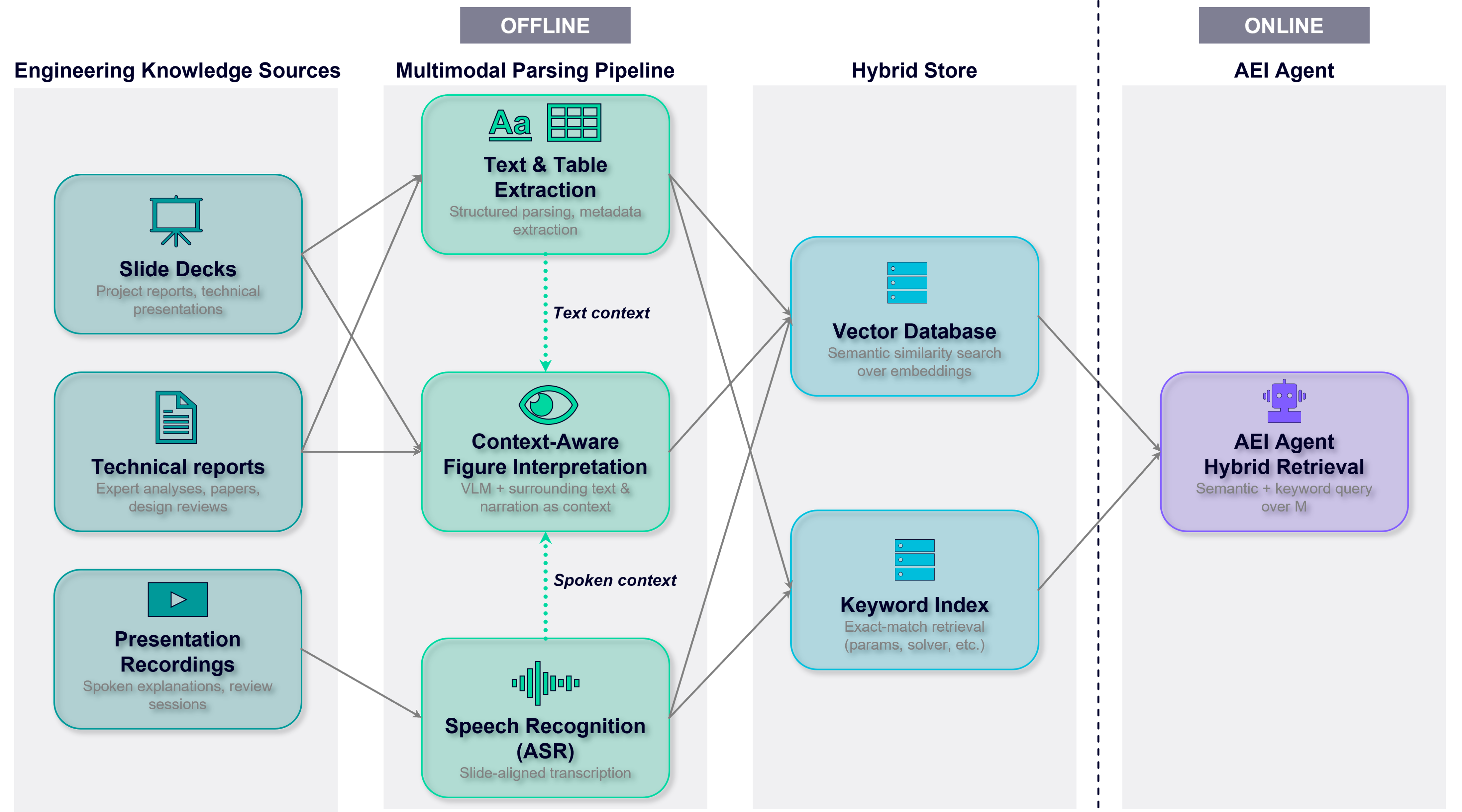}
\caption{Multimodal engineering knowledge processing pipeline for building the offline memory store $D$.
Heterogeneous sources, including slide decks, technical reports, and presentation recordings, are processed through three complementary paths.
The dashed arrows highlight the key mechanism:
surrounding text and spoken narration are supplied as interpretive context to the vision language model (VLM),
enabling physically grounded descriptions of engineering figures.
Parsed artifacts are indexed in a hybrid store that combines semantic and keyword-based retrieval,
which the AEI agent queries during online decision making.}
\label{fig:multimodal}
\end{figure*}

This section presents the AEI formulation consistent with the paper's industrial vision scope. 
AEI provides a common abstraction for workflow state, memory, constraints, and intervention selection across heterogeneous engineering settings.

An engineering workflow is modeled as
\begin{equation}
\mathcal{W} = (\mathcal{S}, \mathcal{O}, \mathcal{A}, \mathcal{Z}, \mathcal{C}, T, U),
\end{equation}
where $\mathcal{S}$ denotes workflow states, $\mathcal{O}$ observations, $\mathcal{A}$ candidate interventions, $\mathcal{Z}$ retrieved workflow memory, 
$\mathcal{C}$ feasibility and specification constraints, $T$ workflow transition dynamics, and $U$ a utility model. 
At step $t$, the latent state $s_t \in \mathcal{S}$ may include artifacts, requirements, simulation outputs, diagnostic logs, tool status, 
and interaction history. Because workflows are only partially observed, the agent acts on an observation $o_t \in \mathcal{O}$ built from available tool outputs, 
reports, status indicators, and retrieved context.

Retrieved workflow memory is represented as
\begin{equation}
z_t = \mathcal{M}(q_t, H_t, D),
\end{equation}
where $q_t$ is a context dependent query, $H_t$ denotes workflow history, and $D$ is an offline memory store built from prior artifacts, 
workflow traces, and human feedback. 
Depending on the application, $z_t$ may contain similar cases, diagnoses, parameter sensitivities, requirement traces, or prior interventions. 
In practice, the broader utility model $U$ may be approximated through a workflow-energy heuristic
\begin{equation}
E_t(a) = \alpha \hat{G}_t(a) + \beta \hat{V}_t(a) + \gamma \hat{C}_t(a),
\end{equation}
where $\hat{G}_t(a)$ estimates the remaining performance gap after intervention $a$, $\hat{V}_t(a)$ estimates residual constraint violation, and $\hat{C}_t(a)$ estimates workflow cost. Lower values indicate more favorable next steps under the current predictive model. This quantity is used as a stability-inspired guidance signal rather than as a formal proof of stability.

A recommendation-centered instantiation of AEI ranks feasible actions using predicted workflow energy:
\begin{equation}
a_t^{*} = \arg\min_{a \in \mathcal{A}^{\mathrm{feas}}_t} E_t(a).
\end{equation}
This rule should be interpreted as an operational heuristic, not as a claim of fully autonomous engineering control. 
When evidence is insufficient, the agent may instead trigger additional retrieval, targeted simulation, or human escalation. 
The remaining sections instantiate this formulation across multimodal knowledge reuse, suspension design, reinforcement learning tuning, 
aerodynamic exploration, and MBSE.

%\FloatBarrier

\section{Learning from Multimodal Engineering Data}

Engineering organizations routinely generate large volumes of knowledge artifacts, including slide decks that summarize simulation campaigns and design decisions, 
technical reports that document analyses and failure diagnoses, and recorded presentations in which engineers explain their reasoning during project reviews.
Taken together, these artifacts encode substantial institutional knowledge, yet they are difficult to reuse systematically because they are distributed across 
shared platforms and embedded in formats that resist simple text search.
As a result, engineers often spend time rediscovering solutions that were already documented in earlier projects, 
while lessons learned are gradually lost as teams evolve over time.

Within AEI, this challenge corresponds to the offline construction of the memory store $D$, which later supports retrieval of workflow memory through $z_t = \mathcal{M}(q_t, H_t, D)$.
Figure~\ref{fig:learning_phases} introduced this offline phase at a conceptual level; here we specify a multimodal processing pipeline that 
transforms real engineering artifacts into reusable memory.
The objective is not merely document retrieval, but preservation of the links among text, figures, speech, and workflow history that make prior engineering 
decisions reusable in subsequent workflows.

A conventional retrieval-augmented generation (RAG) pipeline~\cite{agenticrag2025}, in which documents are divided into text chunks, embedded, 
and retrieved by vector similarity, is inadequate for engineering content for three reasons.
First, much of the critical information resides in figures, including simulation plots, architecture diagrams, and annotated CAD screenshots.
When extracted in isolation, such figures often lack the physical context needed for meaningful interpretation, because their significance 
depends on boundary conditions, solver settings, design intent, or modeling assumptions described elsewhere in the document or presentation.
Recent work on visual document retrieval~\cite{colpali2024} shows that vision language models (VLMs) can embed document images directly, 
avoiding brittle OCR pipelines, but page-level retrieval alone does not capture the cross-page narrative structure typical of engineering presentations.
Second, a substantial portion of engineering rationale is communicated orally during reviews and presentations but never written on the slides themselves.
Design trade-offs, known pitfalls, and informal recommendations therefore remain invisible to document-only pipelines.
Third, engineering retrieval often depends on exact matches for parameter names, solver versions, or model identifiers, 
which pure semantic search handles unreliably.
For AEI, the central issue is not simply whether related content can be retrieved, but whether it is retrieved with sufficient 
engineering context to support the next workflow intervention.

To address these limitations, we construct $D$ through the three-path multimodal parsing pipeline shown in Figure~\ref{fig:multimodal}.

The first path performs \emph{structured text and table extraction} from slide decks and PDF reports.
Text blocks, tables, and metadata such as author, date, and project identifier are parsed and indexed to support both semantic retrieval and exact-match lookup.

The second path performs \emph{context-aware figure interpretation}.
Rather than passing each figure to a VLM in isolation, the system provides surrounding textual context as part of the prompt, 
including slide text, adjacent report passages, and, critically, the time-aligned spoken narration produced by the third path.
This enables the model to generate descriptions that reflect the engineering meaning of the figure rather than only its visual appearance.
For example, a contour plot can be described as indicating thermal concentration near an inlet boundary under a specific operating condition, 
rather than merely as a region of high intensity.
This distinction is essential because the same visual pattern may indicate a design flaw in one setting and expected behavior in another.

The third path applies \emph{automatic speech recognition} (ASR) to presentation recordings.
The resulting transcripts are aligned to individual slides by timestamp, so that spoken explanations become associated with the visual 
and textual artifacts they describe.
This captures rationale and tacit knowledge that exists only in oral commentary, 
such as why a particular mesh density was selected or which failure mode a design modification was intended to prevent.

The outputs of all three paths are stored in a hybrid retrieval system that combines a vector database for semantic similarity search with 
a keyword-indexed database for exact-match queries~\cite{mrag2025}.
This design supports both conceptual queries, such as ``thermal issues in rear diffuser designs,'' and precise technical queries, 
such as ``Star-CCM+ polyhedral mesh, k-$\omega$ SST, 12-degree diffuser angle.''
During online operation, the AEI agent issues hybrid queries against this store as part of its reasoning cycle.
For example, when preparing a simulation for a new vehicle variant, the agent may first retrieve reports from geometrically similar prior projects 
through semantic search.
It may then narrow those results to a specific solver version through keyword filtering.
It can further surface the associated oral explanation in which an engineer discussed convergence difficulties for that class of problems.
Retrieval therefore does not merely return documents; it informs the next workflow action, such as which solver setup to reuse, which failure mode to check first, 
or when to escalate for human review.

Within the overall AEI architecture, this section addresses the offline layer that enables the subsequent online workflows.
The memory constructed here allows the agent to ground future design and analysis decisions in prior cases rather than relying only on the current prompt 
or immediate simulation outputs.
Multimodal engineering memory is therefore not an auxiliary feature of AEI, but a primary mechanism through which institutional 
engineering knowledge becomes available for closed-loop planning, intervention selection, and engineer-supervised decision support.

\FloatBarrier
\section{Suspension Design: Target Cascading and Geometry Optimization}

\begin{figure}[!b]
\centering
\includegraphics[width=\columnwidth]{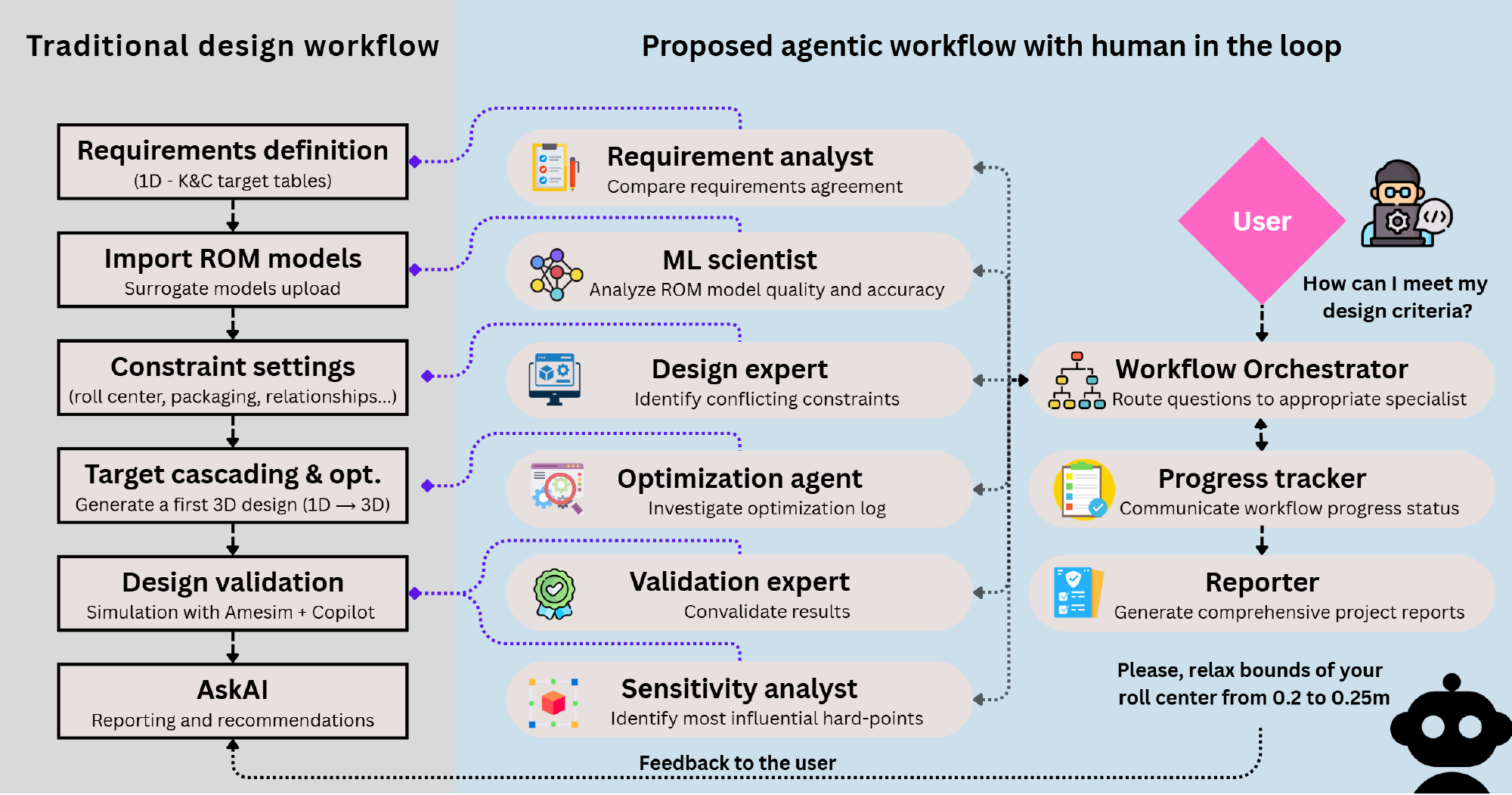}
\caption{Agentic suspension design workflow with human-in-the-loop supervision.
The agent monitors optimization outcomes, active geometric constraints, and prior cases,
and recommends targeted next-step action during hardpoint design.}
\label{fig:suspension}
\end{figure}

Translating suspension performance targets, typically specified as Kinematics \& Compliance (K\&C) curves, into feasible 3D hardpoint configurations 
is a central challenge in early stage vehicle design.
The task involves satisfying multiple coupled constraints, including packaging limits, roll-center placement, scrub radius, and tie-rod inclination, 
while preserving target kinematic behavior.
Because these requirements interact nonlinearly, the design process has traditionally depended on extensive expert iteration, 
diagnostic judgment, and repeated optimization restarts.

Within the AEI framework, this use case emphasizes the online closed-loop decision phase introduced in Figure~\ref{fig:learning_phases}.
At step $t$, the workflow state includes the current hardpoint geometry, target K\&C responses, active geometric and packaging constraints, 
optimization status, and diagnostic traces from prior runs.
From this evolving state, the agent forms an observation of the current design situation, 
retrieves related prior cases or sensitivity information when available, and ranks feasible next-step interventions.
These interventions may include modifying a subset of hardpoints, relaxing or reformulating a constraint, changing the optimization initialization, 
launching an auxiliary sensitivity study, or escalating the case for engineer review.

In a conventional workflow, target cascading and geometry optimization are often treated as a sequential pipeline:
requirements are defined, surrogate or simulation models are prepared, constraints are specified, and optimization is launched.
When the search fails to find a feasible geometry, the engineer must manually inspect logs, infer the likely source of infeasibility, 
revise the setup, and restart from an earlier stage.
As a result, failure acts mainly as a stopping condition, and progress depends heavily on the engineer's ability to diagnose 
intermediate outcomes and choose an appropriate recovery strategy.

The agentic workflow changes this structure by treating each intermediate outcome, including failed optimization runs, 
as an informative workflow state rather than as an endpoint.
Optimization traces, constraint violation patterns, and retrieved memory from prior studies are used jointly to identify 
plausible causes of failure and to prioritize corrective actions.
A run that terminates without a feasible solution may therefore still provide enough information for the agent to recommend 
a productive next step instead of simply triggering manual backtracking.

A representative scenario illustrates the process.
Suppose that the optimizer fails because the roll-center target conflicts with packaging limits and tie-rod geometry.
In a conventional setting, this would appear as an infeasibility report requiring manual diagnosis.
In the AEI workflow, the agent analyzes the violation pattern, retrieves similar prior cases or local sensitivity information, 
and proposes ranked candidate interventions, such as relaxing a bound, adjusting a subset of hardpoints, 
or restarting the search from a more favorable initialization.
The engineer then evaluates these recommendations in the context of broader vehicle-level requirements and 
decides whether to accept, modify, or reject them.

Human-in-the-loop supervision remains central throughout this workflow.
The objective is not to replace engineering judgment, but to make the reasoning around infeasibility, 
trade-offs, and recovery actions more transparent, traceable, and reusable.
For less experienced engineers, the system exposes why a configuration failed and what compromise a proposed intervention implies.
For experienced engineers, it reduces repetitive diagnostic effort and helps focus attention on decisions that materially 
affect vehicle-level performance and feasibility.

\FloatBarrier

\FloatBarrier
\section{Reinforcement Learning Control Tuning}

\begin{figure}[!t]
\centering
\includegraphics[width=\columnwidth]{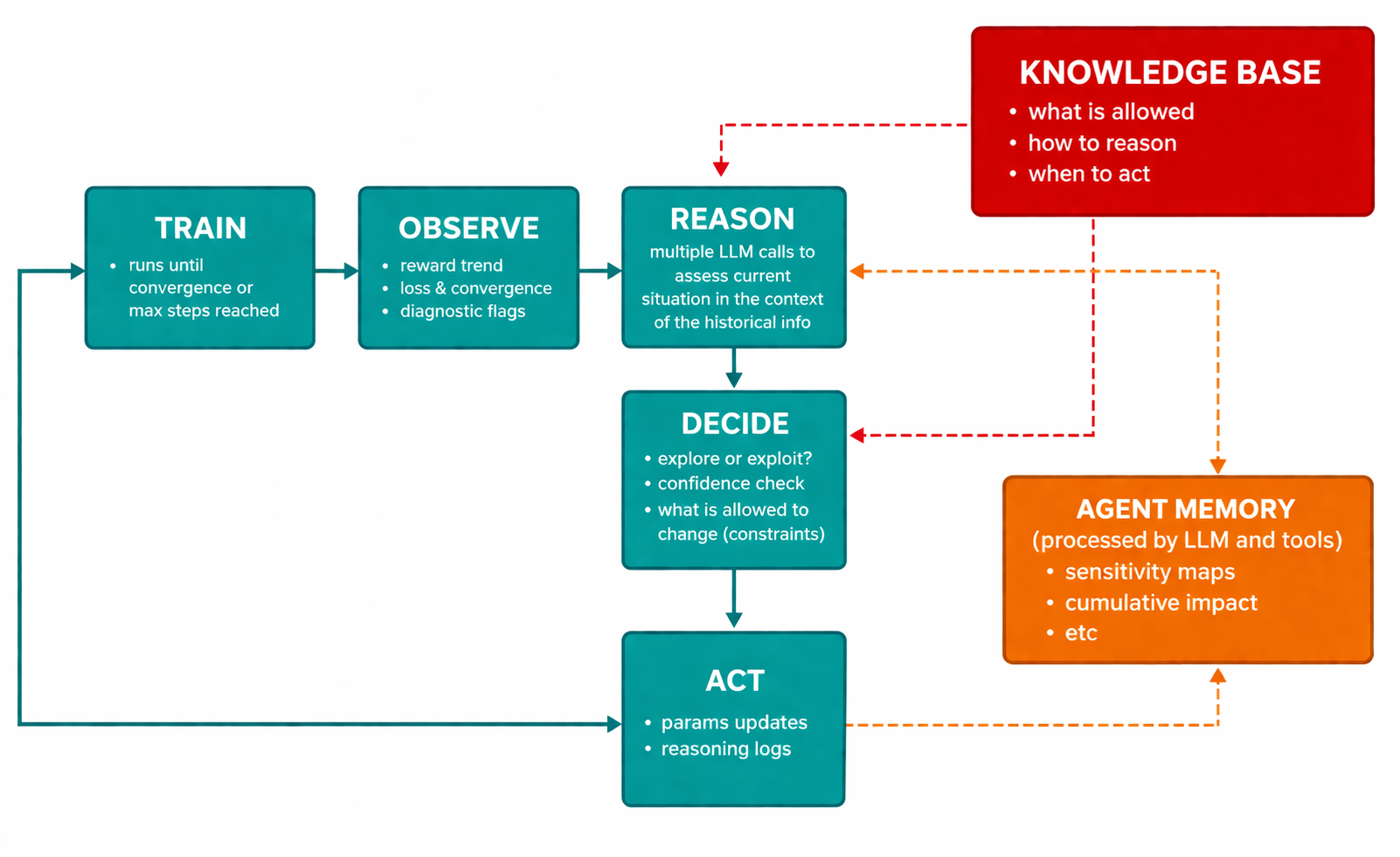}
\caption{Agentic workflow for reinforcement learning hyperparameter tuning.
The agent combines expert knowledge and accumulated run history to interpret training outcomes
and recommend targeted tuning actions.}
\label{fig:rl_optimization_loop}
\end{figure}

Deploying reinforcement learning (RL) for real world control tasks introduces workflow challenges that differ substantially from 
those encountered in simulation-only studies or conventional optimization settings.
In addition to environment design and reward formulation, practitioners must tune a large set of interdependent hyperparameters, such as learning rate, 
entropy regularization, reward weighting, discount factor, and episode horizon, whose effects on training dynamics are nonlinear, 
context-dependent, and difficult to predict in advance.
In practice, much of this knowledge remains tacit, residing with a small number of experienced practitioners, 
which creates steep onboarding barriers and makes tuning workflows difficult to scale and reproduce.

Within the AEI framework, this use case is another instance of the online closed-loop decision phase shown in Figure~\ref{fig:learning_phases}.
At step $t$, the workflow state includes the current controller or policy configuration, training hyperparameters, reward trajectories, 
stability indicators, diagnostic summaries, and the history of prior tuning attempts.
The agent does not act directly on the controlled plant; instead, it intervenes on the engineering workflow that produces the controller.
Given the current workflow state, it retrieves relevant prior tuning episodes or diagnostic patterns when available and ranks 
candidate next-step interventions, such as adjusting selected hyperparameters, revising the search direction, 
requesting additional evaluation, or escalating the case for engineer review.

To support this process, the workflow combines two complementary knowledge sources, as illustrated in Figure~\ref{fig:rl_optimization_loop}.
The first is a human-authored knowledge base that externalizes expert intuitions, diagnostic heuristics, 
and parameter-sensitivity rules that would otherwise remain implicit.
The second is an agent-built memory accumulated through successive training studies, 
in which the agent records parameter changes, observed reward and stability responses, inferred sensitivity patterns, 
and the reasoning associated with each intervention.
Together, these sources provide the historical context needed for workflow-level decision support: expert priors guide interpretation, 
while empirical run history grounds recommendations in observed behavior.

The resulting optimization loop can be viewed as an iterative cycle of observation, reasoning, and intervention.
During an initial exploration phase, the agent perturbs selected hyperparameters to characterize sensitivity, instability, 
and reward responsiveness across the training process.
As evidence accumulates, the workflow shifts toward more targeted interventions that exploit the structured memory built from prior runs.
The role of the agent is therefore not merely to search over parameter combinations, but to interpret observed outcomes 
in light of historical context and recommend the next best engineering action.

A key property of this workflow is traceability.
Each study produces a structured reasoning log that records the hyperparameters applied, the outcomes observed, the prior cases or heuristics consulted, 
and the recommendation that followed.
This supports post-analysis, human review, and reuse across projects.
Compared with conventional hyperparameter search methods, which may identify effective configurations without preserving the rationale behind them, 
the AEI formulation emphasizes an explicit and inspectable record of how tuning decisions were made and how workflow knowledge accumulated over time.
\FloatBarrier

\FloatBarrier

\section{Surrogate-Assisted External Aerodynamic Workflow Design}

\begin{figure}[!t]
\centering
\includegraphics[width=\columnwidth]{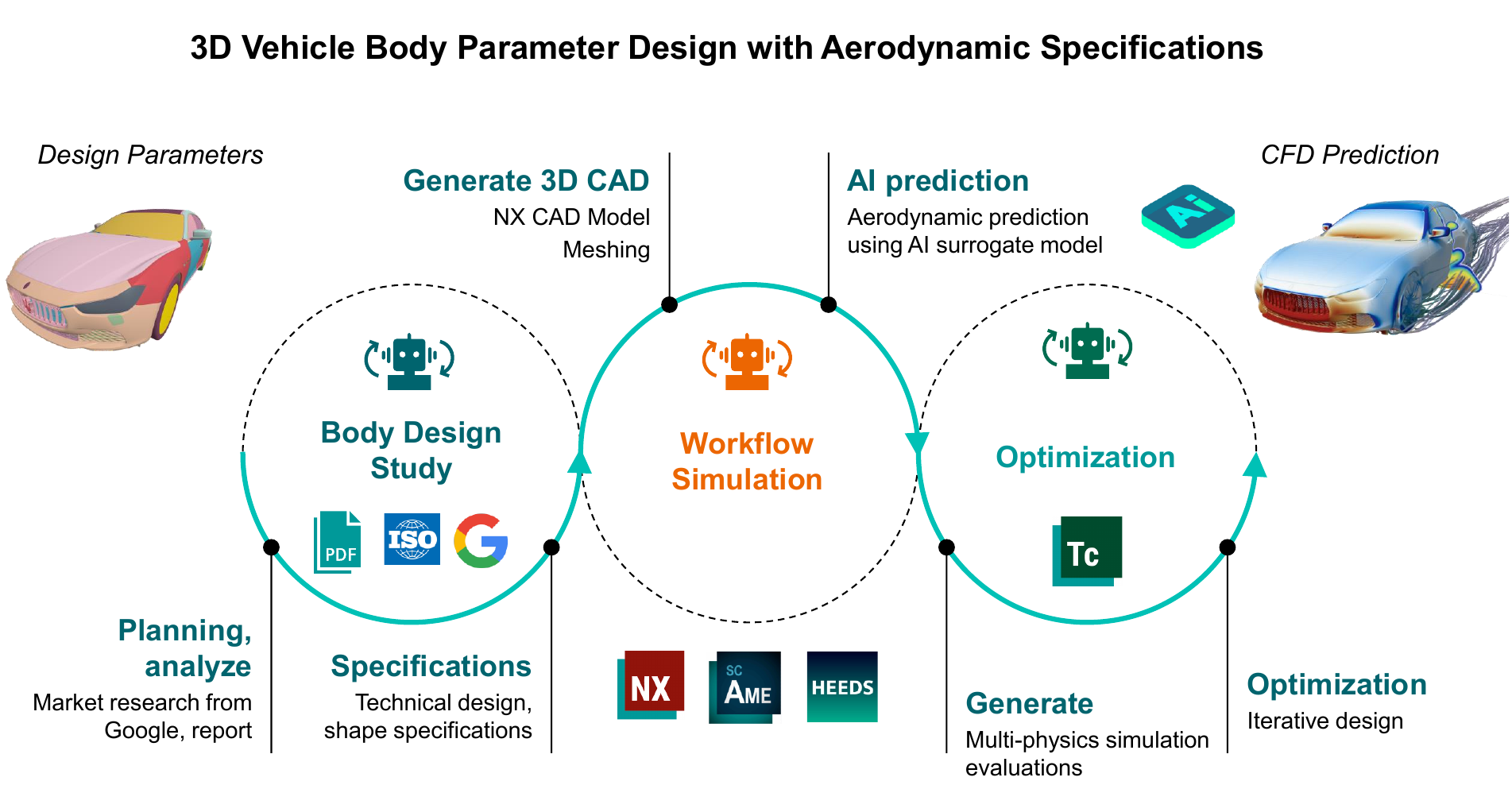}
\caption{Surrogate-assisted external aerodynamic workflow.
The agent uses rapid surrogate predictions to evaluate candidate geometry changes,
reason over performance and feasibility constraints, and decide when higher fidelity CFD analysis is achieved.}
\label{fig:aerodynamic}
\end{figure}

External aerodynamic design in automotive engineering continues to rely heavily on computational fluid dynamics (CFD) for drag reduction, flow control, 
thermal management, and energy-efficiency improvement.
However, high fidelity CFD simulations often require hours per evaluation, which makes broad design exploration expensive when multiple body variants, 
ride heights, operating conditions, and geometric parameters must be assessed.
This challenge is especially significant when engineers must reason over design variables such as vehicle length, mirror position, windshield shape, 
roof profile, or diffuser angle while maintaining aerodynamic performance and feasibility constraints.
To reduce this cost, surrogate models have emerged as a practical means of approximating surface flow fields and integrated aerodynamic metrics with 
sufficient fidelity for early stage engineering decision making.

Recent automotive benchmark efforts have strengthened this landscape.
For example, DrivAerNet family dataset provides a large scale basis for learning and evaluating surrogate models on realistic vehicle geometries 
and high fidelity aerodynamic simulations \cite{drivaernet2024}.
Such benchmarks help establish that learned aerodynamic surrogates can support engineering use, highlighting the growing importance of workflow-level 
coordination of prediction, constraint handling, and higher fidelity data acquisition.

Within the AEI framework, this use case couples the offline and online phases shown in Figure~\ref{fig:learning_phases}.
The offline phase contributes learned surrogate model knowledge, prior design evaluations, and workflow memory from earlier studies.
The online phase uses this information to guide rapid aerodynamic design exploration under target performance and feasibility constraints.
At step $t$, the workflow state includes the current body parameterization, target aerodynamic metrics, surrogate predictions, constraint status, 
and the history of previous design evaluations.

In a conventional CFD-centered workflow, aerodynamic optimization requires repeated solver runs, manual interpretation of trade-offs, 
and careful coordination between geometry changes and physical constraints.
Even when surrogate models are available, they are often used only as standalone predictors rather than as components of a broader decision loop.
AEI instead treats the surrogate as one element of a constrained closed-loop workflow in which the next engineering action is selected in context, 
based not only on predicted aerodynamic response but also on feasibility, prior workflow history, and the expected value of further analysis.

Given target specifications such as a desired drag level and additional design constraints, the agent reasons over the 
parameterized vehicle geometry and proposes candidate design modifications.
The surrogate model then provides rapid estimates of aerodynamic response, enabling the workflow to screen alternatives before committing to 
higher cost simulation.
Based on predicted performance, constraint status, and retrieved prior cases, the agent can rank candidate next-step actions, 
such as accepting a promising design, refining selected parameters, launching additional local optimization, or 
escalating the case to CFD simulation for confirmation.

The key contribution of AEI in this setting is therefore not merely faster aerodynamic prediction, 
but explicit workflow-level reasoning over when and how surrogate-based evidence should drive the next design decision.
In this formulation, the surrogate model accelerates evaluation, while the agentic layer organizes prediction, constraint reasoning, 
and selective fidelity escalation into a traceable engineering process.
\FloatBarrier

\FloatBarrier
\section{Integrating Agentic AI with Simcenter for MBSE Workflows}

\begin{figure}[!b]
\centering
\includegraphics[width=\columnwidth]{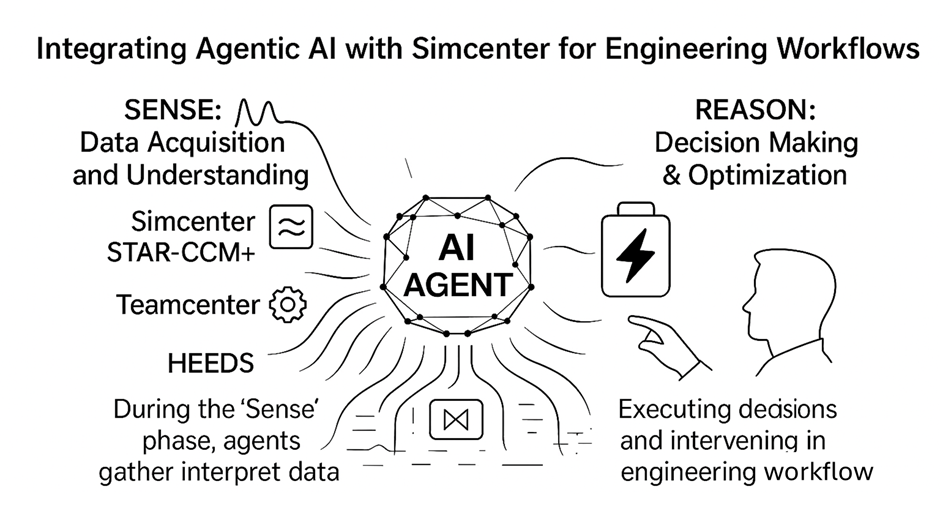}
\caption{Sense - Reason - Act view of Agentic AI integration with the Simcenter engineering toolchain.
The agent estimates workflow state from models, requirements, and analyses, reasons over candidate interventions, 
and supports engineer-supervised workflow execution.}
\label{fig:mbse}
\end{figure}

Model-Based Systems Engineering (MBSE) and Model-Based Design (MBD) provide a natural setting for AEI because they already organize engineering activity 
around explicit models, structured requirements, traceable dependencies, and coordinated toolchains~\cite{lu2022mbseontology}.
In this context, the role of the agent is not limited to answering isolated engineering questions; 
it operates over the engineering workflow itself by connecting requirements, models, simulation outputs, and workflow history to 
support the next engineering decision.
This is not a separate architecture from AEI, but a direct realization of the same closed-loop structure introduced earlier in the paper: 
workflow-state estimation, planning, and control over an engineering toolchain.
Figure~\ref{fig:mbse} illustrates this mapping through a Sense - Reason - Act view, where Sense corresponds to state estimation, 
Reason to planning, and Act to workflow control under engineer supervision.

A representative workflow trace is an electrified vehicle thermal management study.
Suppose that a top-level requirement in Teamcenter is updated so that battery inlet temperature must remain below a specified threshold during fast charging, 
while a cabin cooldown target must also be satisfied under hot-ambient operation.
The current Amesim model satisfies these targets separately, but the combined scenario exposes a conflict: maintaining battery thermal protection requires 
a coolant-flow allocation that degrades cabin cooldown performance and increases auxiliary power demand.
In a conventional workflow, an engineer would manually inspect the changed requirements, identify affected subsystem models, 
determine which studies must be rerun, and diagnose whether the issue originates in control logic, component sizing, or requirement allocation.
AEI instead treats this situation as an explicit engineering state that can be sensed, reasoned about, and acted on systematically.

\noindent\textbf{Sense (workflow-state estimation).}
The agent first estimates the current state of the workflow from the engineering toolchain, including model status, requirement coverage, 
recent analysis outcomes, and indicators of inconsistency or design risk.

In the thermal management trace above, the agent detects that a requirement object in Teamcenter has changed and follows the 
traceability links to the affected system architecture and simulation studies.
It then gathers the latest Amesim results, extracting quantities such as battery inlet temperature, cabin cooldown time, compressor load, 
pump power, and valve states.
If rapid surrogate estimates are available through AI models (i.e. PhysicsAI), the agent can also use them to screen likely thermal consequences 
before committing to a more expensive rerun.
The result is not merely a collection of reports, but a structured state estimate identifying which requirements are violated, 
which subsystems are implicated, and which analyses are reliable enough to inform the next decision.

\noindent\textbf{Reason (planning and intervention selection).}
The agent then combines the current state with workflow memory and engineering constraints to plan candidate interventions, 
reasoning over cross-domain trade-offs, requirement conflicts, and the likely impact of each action.

In the representative thermal management case, the agent determines that the issue is not a single-model failure, but a workflow-level conflict between 
requirement satisfaction and resource allocation.
It consults workflow memory to retrieve comparable past cases, for example studies in which battery protection and cabin comfort 
competed for the same coolant-loop capacity.
Offline learning, preference-informed evaluation, and memory-augmented retrieval help narrow the problem to a small set of plausible next steps.

The agent can then evaluate a limited set of candidate interventions, such as retuning valve and pump control logic, 
reallocating heat-exchanger capacity, modifying a component size, or revisiting requirement decomposition.
The point is not that the agent makes an opaque autonomous decision, but that it performs the planning step of the AEI loop 
by reducing the problem to a small, explainable set of next actions grounded in current evidence and prior workflow history.

\noindent\textbf{Act (workflow control and execution).}
The final step is to execute or propose the selected intervention in the workflow, for example by launching a targeted simulation, 
modifying selected parameters, preparing a requirement review, or escalating a trade-off to a human decision maker.
In this paper, the operating mode is recommendation-centered rather than fully autonomous. The agent organizes candidate interventions, 
but the engineer retains decision authority over requirement changes and other design actions.

Continuing the same trace, the agent may launch a targeted Amesim study with a revised coolant-flow control strategy, request a rapid surrogate-based 
estimate of the resulting thermal margin, and generate a traceability report in Teamcenter showing which requirements, parameters, 
and analyses are affected by the proposed change.
It then presents the engineer with a concise recommendation, for example that the conflict should first be addressed through controller retuning rather 
than component resizing, together with the supporting evidence and the remaining risks.
The engineer retains final authority, but the workflow is accelerated because information gathering, cross-tool reasoning, 
and impact tracing have already been organized into a coherent intervention.
\FloatBarrier

\section{Conclusion}

This paper presented Agentic Engineering Intelligence (AEI), an industrial vision framework for 
modeling engineering workflows as constrained, history-aware sequential decision processes. 
Rather than treating design optimization, diagnosis, control tuning, knowledge reuse, 
and MBSE as isolated tasks, AEI frames them as workflow-level intervention problems supported by memory, 
retrieval, and engineer-supervised decision guidance.

Across representative automotive use cases, the paper showed how diverse workflows can be expressed within 
a common formulation linking offline memory construction with online workflow-state estimation 
and intervention ranking. The proposed workflow-energy heuristic provides a practical, 
stability inspired signal for prioritizing candidate next steps without claiming fully autonomous engineering control.

Overall, AEI positions engineering AI as a problem of process-level intelligence over toolchains, 
artifacts, constraints, and prior cases. 
Future work should focus on empirical validation, benchmarking, uncertainty-aware recommendation, 
and tighter integration with traceable industrial engineering environments.

%\input{ack.tex}

%% ── REFERENCES ────────────────────────────────────────────────

\bibliographystyle{ieeetr}
\bibliography{main}

\end{document}